\documentclass[10pt,twocolumn,letterpaper]{article}

\usepackage{cvpr}
\usepackage{times}
\usepackage{epsfig}
\usepackage{graphicx}
\usepackage{amsmath}
\usepackage{amssymb}
\usepackage{enumitem}

\usepackage{booktabs,tabularx}
\newcolumntype{C}{>{\centering\arraybackslash}X}

\usepackage[linesnumbered,ruled,vlined]{algorithm2e}

\usepackage[breaklinks=true,bookmarks=false]{hyperref}

\cvprfinalcopy 

\def\cvprPaperID{****} 

\begin{document}

\title{Automated Robustness with Adversarial Training as a Post-Processing Step}

\author{Ambrish Rawat, Mathieu Sinn, and Beat Buesser\\
IBM Research Europe\\
IBM Technology Campus, Damastown Ind. Park\\
Dublin, D15HN66, Ireland\\
{\tt\small \{ambrish.rawat, mathsinn, beat.buesser\}@ie.ibm.com}

}

\maketitle

\begin{abstract}

Adversarial training is a computationally expensive task and hence searching for neural network architectures with robustness as the criterion can be challenging. 
As a step towards practical automation, this work explores the efficacy of a simple post processing step in yielding robust deep learning model.
To achieve this, we adopt adversarial training as a post-processing step for optimised network architectures obtained from a neural architecture search algorithm.
Specific policies are adopted for tuning the hyperparameters of the different steps, resulting in a fully automated pipeline for generating adversarially robust deep learning models.
We evidence the usefulness of the proposed pipeline with extensive experimentation across 11 image classification and 9 text classification tasks.

\end{abstract}
\section{Introduction}

The rapidly growing range of Deep Learning applications has created a need for automation of creation and maintenance of neural networks.
Despite their usefulness in many applications, it is well known that deep learning models are vulnerable to adversarial examples which questions their reliability in mission-critical applications \cite{Szegedy2013}.
While the community continues to respond with new approaches to defend models against adversarial attack, the arms race is known to be skewed and very few approaches have been found to be effective for defense. 
Adversarial training~\cite{Goodfellow2015_Explaining,Madry2018_Towards} is among the most promising approaches where a model is trained on adversarial examples.
A key challenge in designing adversarial training protocol is their sensitvity to hyper-parameters.
While the field of Neural Architecture Search (NAS) \cite{Wistuba2019_Survey} has led to the automation of various aspects of deep learning design, it is also a resource intensive process.
Therefore consideration of adversarial robustness as an additional criterion is only going to exacerbate the computational complexity of the search.
We attempt to circumvent this with a simple but effective approach for the automated creation of adversarially robust models by employing Adversarial Training as a post processing step following the search operation. 
The contributions of this work are threefold:
\begin{itemize}
    \item First, it proposes a fully-automated system that, given a dataset and a predictive modelling task, outputs an adverarially robust model. 
    With this work we prescribe design choices and modifications to the adversarial training protocol which generalise its applicability to any dataset.
    \item Second, it demonstrates that effective robustness - as measured by accuracy of the model under bounded $l_\infty$ perturbations - can be obtained at a relatively low cost by incorporating adversarial training as a post-processing step. This is in contrast to searching with the robustness criterion as part of a neural architecture search.
    \item Third, the validity of the approach is supported with an extensive evaluation for 11 image and 9 text classification tasks. 
\end{itemize}

\section{Related Work}

There is a vast body of literature on defenses against adversarial threats, of which Adversarial Training~\cite{Madry2018_Towards} has surfaced as one of teh most promising method.
While the idea of adversarial training is simple and elegant in its min-max objective, it's known to be hard to train with.
Common strategies involve alternate optimisation where first a batch of adversarial examples are computed which are then fed into the regular training objective for optimisation.
This incurs a range of hyper-parameter choices like attacks to be used for computing adversarial examples, the ratio of adversarial samples in the augmented dataset, the frequency of updates etc. 
The popular attack algorithms like Projected Gradient Descent (PGD) themselves rely on a number of hyperparameters like $\epsilon$-budget, step-size and maximum iterations of optimisation. 
The sensitivity of model robustness to these numerous hyperparamters makes it difficult to automate the process of adversarial training.

One of the hallmarks of automation in Deep Learning has been Neural Architecture Search (NAS) \cite{Wistuba2019_Survey}, where the neural architecture, critical to the model's performance, is treated as part of the hyper-parameter search space. 
A natural choice for autumated generation of adversarially robust models is to exntend the NAS algorithms to account for robustness. 
However, most of popular NAS approaches focus on the resulting model accuracy with almost no attention towards adversarial robustness.
Standard protocols for creating adversarially robust models rely on pre-determined hyperparameters of the adversarial examples to be used during the training; also the model architecture is often adopted in ad-hoc fashions to make adversarial training effective~\cite{Madry2018_Towards}. The work in~\cite{Zoph2018} used neural architecture search combined with adversarial training, however, it still used pre-determined hyperparameters and applied ad-hoc modifications to the architectures found during the search. \cite{Sinn2019_Safeml} is the first work which created architectures for adversarially robust models fully automatically; in contrast to our contribution it still resorted to pre-defined hyperparameters.

\section{Automated Generation of Robust Models}

The proposed framework for automated generation of adversarially robust deep learning models comprises of two steps - 
\begin{itemize}[noitemsep]
    \item First, a NAS algorithm is used to obtain an optimal network architecture for the given task.
    \item Second, the obtained architecture is further trained with an adversarial training protocol for hardening it against adversarial attacks.
\end{itemize} 
We describe these two steps in detail in the following sections. 

\subsection{Neural Architecture Search}

Automation  of  Deep  Learning  seeks  to  automate  the different design decisions required for employing a deep learning algorithm.
NAS algorithms choose the optimal architecture by casting the search as a hyperparameter optimisation problem where the model parameters $\theta$ for a deep learning model $m$ are estimated by minimizing a loss function $\mathcal{L}$ with respect to the training data $\mathcal{D}_\text{train}$, and the hyperparameters including the architecture $\alpha$ are learnt by maximising an objective function $\mathcal{O}$ (often same as the negative loss function) on the validation partition $\mathcal{D}_{\text{valid}}$:
\begin{eqnarray}
    \label{eq:dl_opt}
    \lambda\left(\alpha,D\right) &=& \arg \min_{\theta} \mathcal{L}(m_{\alpha,\theta},D_{\text{train}}) \\
    \label{eq:hyp_opt}
    \alpha^\ast &=& \arg \max_{\alpha}\mathcal{O}( \lambda\left(\alpha,D\right), D_\text{valid})
\end{eqnarray}

Common optimizers for NAS are inspired from approaches in reinforcement learning or evolutionary algorithms.
Neuro-Cell-based Evolutionary Search (NCEvolve)~\cite{Wistuba2018_Deep} is an effective NAS algorithm that uses an evolutionary search over a pool of candidate architectures, which we utilise for the first step of our proposed pipeline.


\subsection{Hardening as Post-processing Step (HAPS)}

The second step in our approach, which we call Hardening As Post-processing Step (HAPS), involves the use of Adversarial Training.
Adversarial Training builds on the idea that optimising a deep learning model with respect to an augmented training set with samples perturbed as per a threat model will result in a model that is robust to such perturbations.
As in \cite{Madry2018_Towards}, we permit a set of perturbations $S$ within which the adversary can seek for adversarial examples, leading to the following min-max objective of adversarial training:
\begin{equation}
    \label{eq:adv}
    \min_\theta \mathbb{E}_{(x,y)\sim D_{\text{train}}}\left[\max_{\delta\in S}\mathcal{L}\left(m_{\alpha^\ast,\theta},x+\delta, y\right)\right]
\end{equation}

We use the Project Gradient Descent (PGD) attack~\cite{Madry2018_Towards} to approximate the computation of the inner maximum. 
Given an attack budget specified by maximum allowed perturbation $\epsilon$, a step size $\epsilon_{\text{step}}$ and permitted number of iterations $n$, PGD iteratively computes the perturbation as $\operatorname{sign}\left(\nabla_{\boldsymbol{x}}\mathcal{L}\left(m_{\alpha^\ast,\theta}\left\{x,y\right\}\right)\right)$.
The gradient update in the adversarial training is suitably adopted to include a fraction of adversarial samples ${x}_{\mathrm{adv}}$ along with bening samples $x$ for each minibatch.

Automation of this post-processing step raises several questions: how to chose the attack budget $\epsilon$ for PGD during adversarial training? And what is an appropriate fraction for adversarial images during the minibatch updates?
In order to make HAPS applicable to any dataset, we address these with specific policies based on empirical observations.

There are three notable aspects of HAPS.
First, for a model obtained from NAS, the adversarial training is performed with Stochastic Gradient Descent with cosine-annealing schedule for learning rate with no warm restarts.
Second, the fraction of adversarial images for every minibatch $\nu$ is increased from 0 to 0.5 with a complimentary annealing schedule (Algorithm \ref{algo:haps}). 
Popular choices for the fraction are 0.5 and 1.0 (see \cite{Sinn2019_Safeml}), however we note during our experiments that - without annealing - the adversarial training often collapses for a trained model obtained from NCEvolve for these values.
And finally, the attack budget for PGD specified in terms of $\epsilon$ is also annealed from 0 to $\epsilon_{\text{max}}$.
The exact values for $\epsilon$ for annealing depend on the properties of the datasets.
The sensitivity to perturbations varies across datasets and models; we note that the training is stable when the $\epsilon$ is gradually increased.
For instance, for image datasets with pixel values in range $(0,255)$, we increase the $\epsilon$ from 0 to 16 as powers of 2, and for every $\epsilon$ we train the model for a fixed number of epochs $n_{\text{epochs}}$. 
We fix the maximum iterations $n$ for PGD as 30 during adversarial training and dynamically adopt $\epsilon_{\text{step}}$ as $(1.5 * \epsilon)/n$.
The complete HAPS algorithm is summarised in Algorithm \ref{algo:haps}. 

\begin{algorithm}[h]
\label{algo:haps}
\SetAlgoLined
    \SetKwInOut{Input}{input}
    \SetKwInOut{Output}{output}
    \SetKw{KwBy}{in}
\Input{$x_{\text{train}}, y_{\text{train}}, m_{\alpha^\ast,\theta}$}
\Output{$m_{\alpha^\ast,\theta^\prime}$}
initialise $\eta_{\text{init}}, T, M, n, \nu$\\
 \For{$\epsilon\gets0$ \KwTo $\epsilon_{\text{max}}$}{
    \For{$t$ \KwBy $\{1\dots T\}$}{
    $\gamma\gets \frac{1}{2}\left(1+\cos\left(\frac{t}{T}\pi\right)\right)$ \\
    $\eta\gets \eta_{\text{init}} * \gamma$\\
    $K\gets \lfloor\nu * M * (1-\gamma)\rfloor$\\
    
    Sample a minibatch $(x,y)$ \textup{of size} $M$\\
    $x_\text{adv}\gets$ PGD$(\epsilon,\epsilon_{\text{step}},n, x)$\\
    $\theta \gets \theta - \eta\sum_{i=0}^{K}\nabla_\theta\mathcal{L}(m_{\alpha^\ast,\theta},x_{\text{adv}},y) -\eta\sum_{i=K+1}^{M}\nabla_\theta\mathcal{L}(m_{\alpha^\ast,\theta},x,y)$\\

    }
}
 \caption{Hardening As Post-processing Step}
\end{algorithm}

\section{Experiments}

We validate the effectiveness of our approach by performing an extensive evaluation of the HAPS algorithm. 
As automation is at the centre of this work, we deem it critical to analyse the effect of different hyperparameters and annealing policies.
We examine this for the case of CIFAR10~\cite{cifar} and compare the performance of the obtained model with the state-of-the-art approaches.
We follow this with an experiment where we demonstrate the effectiveness across 11 image- and 9 text- classification tasks. 

\paragraph{Robustness Evaluation}
Across all experiments we adopt the convention of \cite{Madry2018_Towards} for evaluating robustness.
More specifically, we measure the robustness in terms of accuracy for an untargeted PGD attack with maximum $l_\infty$ perturbation of $8$ for images with pixel values in range $(0,255)$, and $0.2$ for GloVe-embedding based text classification models. For the computation of PGD attacks, we use the Adversarial Robustness Toolbox (ART)~\cite{Nicolae2018}.

\paragraph{CIFAR10}
We use the NeuNets\cite{Sood2019_NeuNets} setup for NCEvolve and obtain an architecture for CIFAR10.
This model is then post-processed with HAPS for $46,000$ iterations ($T$) with batch size of $32$, $\epsilon_{\text{max}}$ of 16.0 and maximum adversarial fraction $\nu$ of 0.5. 
We consider the model obtained from NCEvolve as a baseline and compare the accuracy of the models on benign and $l_\infty$ perturbed samples (Table~\ref{tab:cifar}).
The approach in \cite{Sinn2019_Safeml} serves as another baseline where the model is obtained through NCEvolve by modifying the fitness evaluation criterion which however, as discussed in that work, is highly time consuming.
While HAPS achieves a lower accuracy on benign samples, it takes significantly less time than to ~\cite{Sinn2019_Safeml} obtain the model.
We argue that including robustness as an additional search criteria offers sizeable overhead which can be be circumvented with adversarial training conducted as post-processing.
For a similar adversarial robustness, we note a 10\% compromise on benign accuracy with 5x speed-up in overall training.
Although the model trained is different in terms of its architecture, we also compare our model with the one reported in \cite{Madry2018_Towards} (Table~\ref{tab:cifar}).

\begin{table}[]
    \centering
    \caption{Robustness evaluation of different models for CIFAR10}
    \label{tab:cifar}
    \begin{tabularx}{\linewidth}{ c *{4}{c} }
        \toprule
        & NCEvolve & \cite{Sinn2019_Safeml} & \cite{Madry2018_Towards} & HAPS\\
        \midrule
        Acc. &   93.3    &   93.2    &   87.3   &   82.8 \\
        \midrule
         Robust Acc. &  0.0    &   46.3     &   45.8   &   \textbf{46.4}\\
        \midrule
        Time budget &   $\sim$1 day   &  $\sim$14 days     &     &   $\sim$~2 days   \\
        \bottomrule
    \end{tabularx}
\end{table}

\begin{figure}[h]
\begin{center}
   \includegraphics[width=0.6\linewidth]{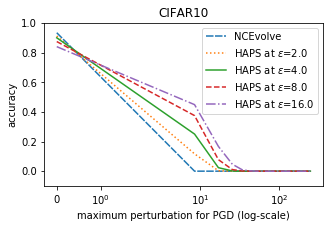}
\end{center}
   \caption{Robustness evaluation for phases of HAPS with PGD - $\epsilon$ ranging from 0 to 255 with $n=200$ and $\epsilon_{\text{step}}=(1.5 * n)/\epsilon$.}
\label{fig:anneal}
\end{figure}

\paragraph{Ablation Study}
We note that optimisation is often unstable when trained model obtained from NCEvolve is post processed with fixed $\nu$ of 0.5.
Similarly, choosing an appropriate $\epsilon$ for PGD is a challenging task. 
We argue that such value of $\epsilon$ is a property of the dataset.
For many datasets, we observe that directly training for large $\epsilon$ values like $16.0$ leads to unstable optimisation. 
We therefore, choose an annealing policy where we gradually increase the $\epsilon$ as $\{1,2,4,8,16\}$ for images and $\{0.1,0.2,0.4,0.8\}$ for text. 
Figure \ref{fig:anneal} shows the effect of this on robustness.

\paragraph{Image and Text Classification Tasks}
We use NCEvolve as set up in NeuNets~\cite{Sood2019_NeuNets} to obtain models for 11 datasets and post process them with HAPS (Table \ref{tab:image}).
Furthermore, in the context of this work we use reduced versions of the datasets with dimensions 32x32.

The NeuNets implementation of NCEvolve for text classification uses 200 dimensional GloVe-embeddings~\cite{Pennington2014_Glove} as a preprocessing step.
We keep the preprocessing in our setup and perform adversarial training for the model that follows the embedding layer.
Similar to the work in \cite{Miyato2016} we compute adversarial attacks with respect to the embedding and report the results in Table \ref{tab:text}.

For both large-scale experiments, we note across the board that HAPS results in improvement of accuracy on perturbed images without significant loss to accuracy on benign samples.
In fact for some datasets we note an improvement in accuracy on benign samples, for instance Rotten-Tomatoes \cite{rotten-tomatoes} (Table \ref{tab:text}) which is in alignment with observations made in \cite{Miyato2016}.
We observed that the annealing of $\epsilon$ was critical for the case of SNIPS~\cite{snips}, where the training almost always collapsed for larger values of $\epsilon$. 
The datasets of Food101~\cite{food101} and Flowers102~\cite{flowers} were found to be challenging where HAPS didn't lead to significant improvements in robustness.
Both these datasets have a low ratio of number of images per class in the training set which we speculate results in the poor mixing during adversarial training. 


\begin{table}[h]
    \centering
    \caption{Robustness evaluation for image classification}
    \label{tab:image}
    \begin{tabularx}{\linewidth}{ c *{4}{C} }
        \toprule
        & \multicolumn{2}{c}{NC Evolve} & \multicolumn{2}{c}{HAPS} \\
        \midrule
            Dataset    & Acc & R.Acc & Acc & R.Acc \\
            \midrule
        F-MNIST \cite{fashion}      &   94.95   &   0.03    &   93.30   &   87.19   \\
        CIFAR10 \cite{cifar}        &   93.3    &   0.00    &   82.87   &   46.42   \\
        CIFAR100 \cite{cifar}       &   65.44   &   2.29    &   51.43   &   16.80   \\
        GTSRB \cite{gtsrb}          &   99.91   &   40.8    &   99.51   &   68.50   \\
        STL10 \cite{stl10}          &   84.24   &   0.51    &   56.97   &   31.07   \\
        SVHN  \cite{svhn}           &   96.73   &   0.70    &   96.53   &   58.64   \\
        Flowers5 \cite{flowers}     &   78.12   &   3.12    &   74.37   &   50.93   \\
        Flowers102 \cite{flowers}   &   54.26   &   4.53    &   45.17   &   13.91    \\
        Quickdraw \cite{quickdraw}  &   65.61   &   5.79    &   65.12   &   56.28   \\
        Caltech256 \cite{caltech}   &   39.22   &   0.43    &   38.19   &   16.98   \\
        Food101  \cite{food101}     &   42.51   &   0.75    &   29.82   &   6.15    \\
        \bottomrule
    \end{tabularx}
\end{table}

\begin{table}[h]
    \centering
    \caption{Robustness evaluation for text classification}
    \label{tab:text}
    \begin{tabularx}{\linewidth}{ c *{4}{C} }
    \toprule
    & \multicolumn{2}{c}{NC Evolve} & \multicolumn{2}{c}{HAPS}\\
    \midrule
    Dataset    & Acc & R.Acc & Acc & R.Acc \\
        \midrule
    CoLA  \cite{cola}                       &   70.52   &   1.89    &   70.40   &   70.46  \\
    Yelp \cite{yelp}                                    &   56.54   &   0.0   &   41.82   &   11.08     \\
    Stanford \cite{stanford}                &   60.17   &   0.0     &   63.04   &   49.57 \\
    IMDB  \cite{imdb}                       &   82.66   &   0.0     &   74.55   &   26.51 \\
    SMS-spam     \cite{sms}                 &   98.54   &   0.0     &   98.66   &   86.38 \\
    SNIPS   \cite{snips}                    &   98.43   &   0.0     &   98.43   &   57.81     \\
    Rotten-Tomatoes \cite{rotten-tomatoes}  &   59.31   &   0.0     &   63.52   &   50.89   \\
    TREC  \cite{trec}                                  &   85.20   &   0.0     &   84.68   &   12.70     \\
    YouTube  \cite{youtube}                              &   93.05   &   0.0     &   94.09   &   44.79   \\
    \bottomrule
    \end{tabularx}
\end{table}

\section{Conclusions}
We demonstrated an end-to-end pipeline for the automatic creation of adversarially robust deep learning models.
To the best of our knowledge, there is no previous work that was able to create such models across a similar variety of datasets.
Our work highlights the benefits of using adversarial training as a post-processing step over using it as an additional search criterion in NAS approaches.
We show that the former achieves similar adversarial robustness with a sizeable speedup in overall computation time.
As a next step it would be interesting to explore how to further optimise the model hardening pipeline and investigate other dimensions like - how to adopt the pipeline to broader threat models besides $l_\infty$ bounded perturbations, and how do models from other NAS approaches influence the performance of the pipeline.

{\footnotesize
\bibliographystyle{ieee_fullname.bst}
\bibliography{egbib}
}

\end{document}